\documentclass{article}
\PassOptionsToPackage{numbers, compress}{natbib}

\usepackage[final]{neurips_2022}
\usepackage[utf8]{inputenc} 
\usepackage[T1]{fontenc}    
\usepackage{hyperref}       
\usepackage{url}            
\usepackage{booktabs}       
\usepackage{amsfonts}       
\usepackage{nicefrac}       
\usepackage{microtype}      
\usepackage{xcolor}         

\usepackage{amsmath,amssymb,amsfonts}
\usepackage{algorithmic}
\usepackage{graphicx}
\usepackage{subcaption}
\usepackage{booktabs, multirow}
\usepackage{changepage,threeparttable}
\usepackage{textcomp}
\usepackage{xcolor}
        
\title{TPFNet: A Novel \underline{T}ext In-\underline{p}ainting Trans\underline{f}ormer for Text Removal}

\author{%
  Onkar Susladkar\thanks{Email: onkarsus13@gmail.com} \\
    Independent Researcher\\
  \And
  Dhruv Makwana \\
  Independent Researcher \\  
  \AND
  Gayatri Deshmukh  \\
  Independent Researcher \\
  \And
  Sparsh Mittal \\  
  IIT Roorkee \\
  \And
  Sai Chandra Teja R \\  
  Independent Researcher \\ 
  \And
  Rekha Singhal \\
  TCS Research, India
}

\begin{document}

\maketitle

\newcommand{\ApproxSign}{\raise.17ex\hbox{$\scriptstyle\sim$}}
\vspace{-0.5cm}

\begin{abstract}
 Text erasure from an image is helpful for various tasks such as image editing and privacy preservation. In this paper, we present TPFNet, a novel one-stage (end-to-end) network for text removal from images. Our network has two parts: feature synthesis and image generation. Since noise can be more effectively removed from low-resolution images, part 1 operates on low-resolution images. The output of part 1 is a low-resolution text-free image. Part 2 uses the features learned in part 1 to predict a high-resolution text-free image. In part 1, we use ``pyramidal vision transformer'' (PVT) as the encoder. Further, we use a novel multi-headed decoder that generates a high-pass filtered image and a segmentation map, in addition to a text-free image. The segmentation branch helps locate the text precisely, and the high-pass branch helps in learning the image structure. To precisely locate the text, TPFNet  employs an adversarial loss that is conditional on the segmentation map rather than the input image. On Oxford, SCUT, and SCUT-EnsText datasets, our network outperforms recently proposed networks on nearly all the metrics. For example, on SCUT-EnsText dataset, TPFNet has a PSNR (higher is better) of 39.0 and text-detection precision (lower is better) of 21.1, compared to the best previous technique, which has a PSNR of 32.3 and precision of 53.2. The source code can be obtained from \url{https://github.com/CandleLabAI/TPFNet}
\end{abstract}

\section{Introduction}
\label{sec:intro}

Recent years have seen phenomenal growth in text recognition from images \cite{jung2004text,patel2012optical,singh2013optical}. Real-life images also contain private or sensitive information such as addresses, cellphone numbers, and other personally identifiable information. Automatic text recognition engines can collect this information and use it for malicious purposes such as marketing, privacy breaches, and identity theft. Text erasure refers to erasing only the text area in the image without changing the pixel values of other areas in the image. Thus, text erasure is useful for many applications, such as privacy protection, autonomous driving, support systems for the visually impaired, information provision systems in exhibition halls, and translation guidance systems for foreigners. For example, after erasing an advertisement text in one language (e.g., English), the text in another language (e.g., Japanese) can be inserted.

However, erasing the text from images presents unique challenges. Text in images does not usually have any border lines or clear distinction from the background. It is often difficult to utilise border or background colour information to detect the text area. The border detection may fail if the background colour and text colour are the same. A technique is needed to recognise the text at all the images' locations. If the image is photographed at an inclined angle, overlapping adjacent text characters may make text recognition difficult. Further, text erasure techniques must deal with background, texture, format, lighting conditions, font, and layout variations.

Tursun et. al. \cite{tursun2019mtrnet} uses a two-stage approach, first ascertaining the text location and then using a segmented mask for removing the text. However, their efficacy depends on the text-detection step. Furthermore, due to the need to train two separate networks, they require a complex training strategy. Some text extraction techniques first recognise individual characters and then connect them to form words. These techniques assume that the character sizes are uniform or make other assumptions about text position. Recent works (e.g., \cite{zhang2019ensnet}) have focused on end-to-end text erasure, where the system takes an input image and produces a final text-free image.

\textbf{Contributions:} In this paper, we propose TPFNet, a deep-learning model for end-to-end text erasure. As such, TPFNet falls into the category of one-stage text removal networks. Our contributions can be summarised as follows:

1. We note that it is simpler to eliminate noise from a low-resolution image than from a high-resolution one. Based on this, the TPFNet network has two parts: feature synthesis and image generation. In part 1, the network learns the features and creates a low-resolution text-free image, and in part 2, the network uses the learned features to predict a high-resolution text-free image.

2. The generator in part 1 uses a pretrained Pyramidal Vision Transformer (PVT) as the encoder and a novel multi-branch decoder. Specifically, the decoder has three branches that predict a high-pass filtered image, a segmentation map, and an image devoid of text. Each of these outputs assists the model in producing accurate results. Specifically, the high-pass filter branch helps in learning and regaining the structural knowledge of the image. The segmentation branch helps in precisely locating the text.

3. Unlike other conditional GAN architectures, TPFNet employs an adversarial loss conditional on the segmentation map rather than the input image. This allows TPFNet to pinpoint the precise location of text within an image (Section \ref{sec:lossFunctionG1}).

4. We rigorously evaluate our network on the Oxford, SCUT, and SCUT-EnsText datasets. On the SCUT and SCUT-EnsText datasets, TPFNet outperforms previous work on all the metrics, and on the Oxford dataset, it outperforms previous on all the metrics except precision of text-detection.
Qualitative results confirm these findings, and the ablation results provide further insights into the importance of each component. With a PVT backbone, TPFNet has 59.8M parameters and a throughput of 14 frames per second. By changing the backbone, a tradeoff can be exercised between image quality and throughput/model-size.

\section{Related Work}
\label{sec:related_work}

Text erasure requires two subtasks: (1) positioning of the text area and (2) erasing the text. Zhang et al. \cite{zhang2019ensnet} propose an end-to-end word erasure network, which combines two subtasks. The discriminator and various loss functions guide the learning of the generator.
In order to enable the network to perceive the text location better, Liu et al. \cite{liu2020erasenet} further introduce a mask branch for learning.
Tursun et al. \cite{tursun2019mtrnet} introduce text segmentation results in the input so that the network can more accurately perceive the position information of the text area. Bian et al. \cite{bian2022scene} achieve specific text stroke detection and stroke removal through a cascading structure. However, these methods need to know the exact location of the text area in advance. Tursun et al. \cite{tursun2020mtrnet++} introduce a fine-tuning sub-network to reduce the network's dependence on the input location information, thereby achieving a more robust text erasure algorithm.

SceneTextEraser is a patch-based auto-encoder featuring skip connections proposed by Nakamura et al. \cite{nakamura2017scene}.  EnsNet \cite{zhang2019ensnet} uses a local-aware discriminator and four improved losses to keep the erased text uniform. Liu et al. \cite{liu2020erasenet}  introduce EraseNet, which has a  coarse-to-fine architecture and an extra segmentation head to assist with text localization. Compared to EraseNet, MTRNet++ \cite{tursun2020mtrnet++} splits the encoding of the image content and the text mask into two distinct streams. Liu et al. \cite{liu2022don} develop a Local-global Content Modeling (LGCM) block using CNNs and Transformer-Encoder to construct long-term relationships among pixels; this improves efficacy of text removal.

\section{Dataset}
\label{sec:dataset}

We use the following three open-source datasets for experimental purposes: All three datasets have images of size $512\times512$.

\textbf{1. The Oxford Synthetic Real Scene Text Detection \cite{gupta2016synthetic} dataset:} This dataset includes \ApproxSign800,000 synthetic images. We picked 95\% of the data for training, 10,000 images for testing, and the rest for validation. Characters, words, and text lines are all annotated in great detail in the dataset. They are made by combining natural images with text produced in a variety of fonts, sizes, orientations, and colors. To provide a realistic appearance, the text is generated and aligned to carefully selected image areas.

\textbf{2. SCUT Synthetic Text Removal \cite{zhang2019ensnet} dataset:} This dataset has 8000 training and 800 test images.

\textbf{3. SCUT-EnsText \cite{6628859} dataset:} The SCUT-EnsText has 3562 images with various text properties. We randomly choose around 70\% of the images for training and the remaining images for testing to guarantee that they have the same data distribution. The training set includes 2749 images and 16460 words, while the testing set has 813 images and 4864 words.

\section{TPFNet: Our Proposed Network}
\label{sec:proposed_method}
\begin{figure}[htbp]
  \centerline{\includegraphics [width=0.95\textwidth]{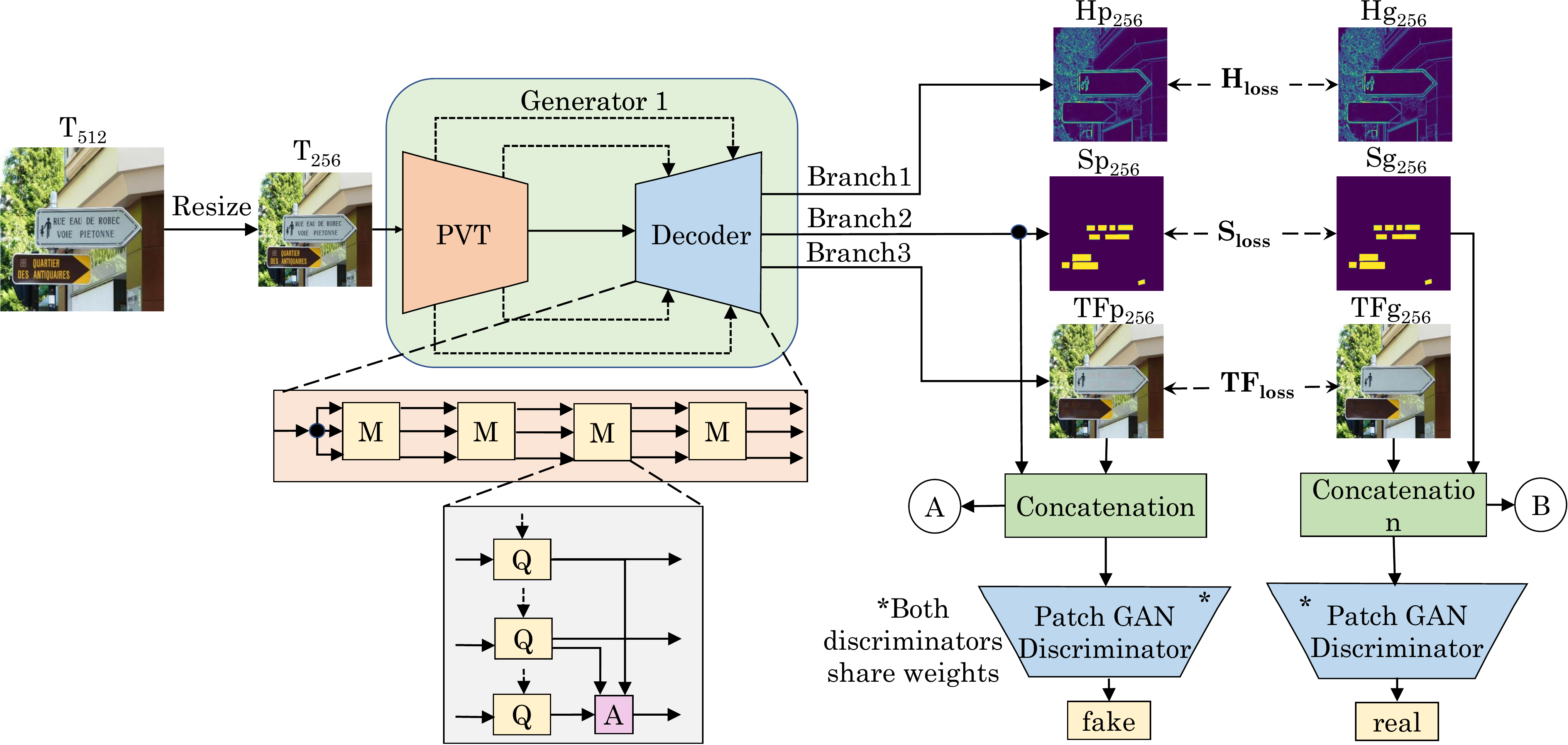}}
  \caption{Feature synthesis (part 1) in TPFNet. (\textcircled{\raisebox{-0.9pt}{A}} and \textcircled{\raisebox{-0.9pt}{B}} connect with Part 2 (Figure \ref{fig:stage_2})) }
  \label{fig:stage_1}
\end{figure}

\textbf{Overall idea:} Our approach works by obliterating the text and painting a reasonable substitution in its place. We utilise the key idea that producing the features from a low-dimensional image is significantly simpler than producing them from a high-dimensional image.
Based on this, our network operates in two parts, and the design of these two parts is shown in Figures \ref{fig:stage_1} and \ref{fig:stage_2}, respectively. The $512\times512$ input image is first resized to create a $256\times256$ image ($T_{256}$), which is fed to part 1. 
Part 1 of TPFNet builds a $256\times256$ text-free image. Part-2 of TPFNet uses this image to create a $512\times512$ text-free image. In essence, part-1 synthesizes various features, and part-2 decodes those features into a high-quality image.

We now describe the architecture and training methodology of feature synthesis (part-1) in Sections \ref{sec:encoder} through \ref{sec:lossFunctionG1} and that of image generation (part-2) in Section \ref{sec:part2}.

\subsection{Encoder design}\label{sec:encoder}
As mentioned above, the part-1 generator receives a $256\times256$ image. The Part 1 generator uses an encoder-decoder architecture. For the encoder, we use the pyramidal vision transformer (PVT) \cite{wang2021pyramid} pretrained on the ImageNet 2012 dataset. Although PVT has higher parameter-count than some other backbone networks (refer Section \ref{sec:ablation}), it provides unique benefits. PVT may be trained on dense image partitions to produce high output resolution, which is critical for dense prediction. PVT also uses a gradual shrinking pyramid to minimise calculations for big feature maps. These characteristics set PVT apart from ViT (vision-transformer), which often produces low-resolution outputs while incurring large computational and memory overheads. PVT extends the transformer framework with a pyramid structure, allowing it to produce multi-scale feature maps for dense prediction applications. As a result, it brings together the advantages of both CNN and the transformer.
Our model uses multi-scale feature maps learned by PVT to acquire the representational knowledge needed to successfully eliminate text. The output from the final part of PVT is fed into the generator's decoder, which reconstructs the images.

\subsection{Decoder design}
\textbf{Key idea:} In order to maximise the network's learning potential, we employ a multi-headed decoder structure in the generator and train it to simultaneously predict segmentation maps, high-pass filtered images, and text-free images. Our decoder integrates three branches' characteristics, providing three outcomes: a high-pass filtered image, a binary segmentation map, and a text-free image. The segmentation branch ascertains the text position, and the high-pass filter branch extracts the edges to obtain the object structure. The attention block fuses the features learned by the segmentation branch and high-pass-filter branch with those learned by the text-free image generation branch. The use of three branches allows every branch to learn to forecast its own output better and to aid the other two branches by supplying extra learned representations to them. We use the Laplacian filter as the high-pass filter.

\textbf{Architecture:} The decoder has four identical modules, shown as ``M'' in Fig. \ref{fig:stage_1}. Module ``M'' has three Q-blocks and one attention-block, and they are designed as follows.

\begin{figure}[htbp]
  \centerline{\includegraphics [width=0.9\textwidth]{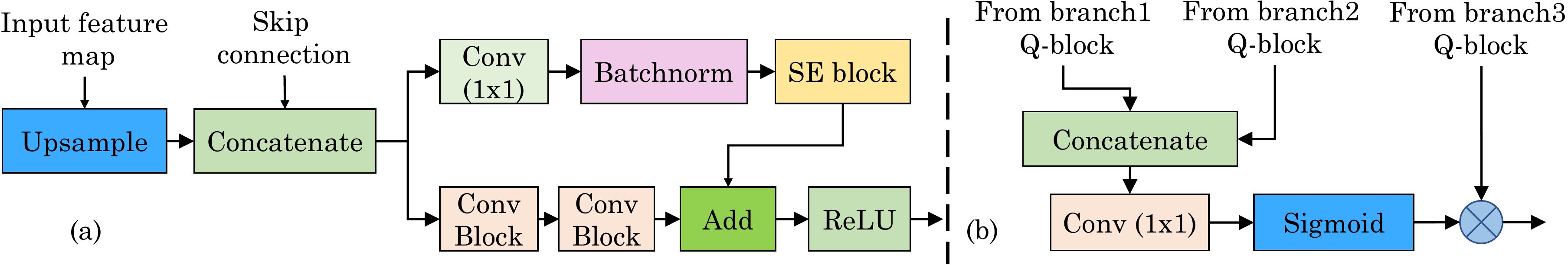}}
  \caption{(a) Q-Block, (b) Attention-Block}
  \label{fig:ad_block}
\end{figure}

\begin{figure}[htbp]
 \centerline{\includegraphics [width=0.95\textwidth]{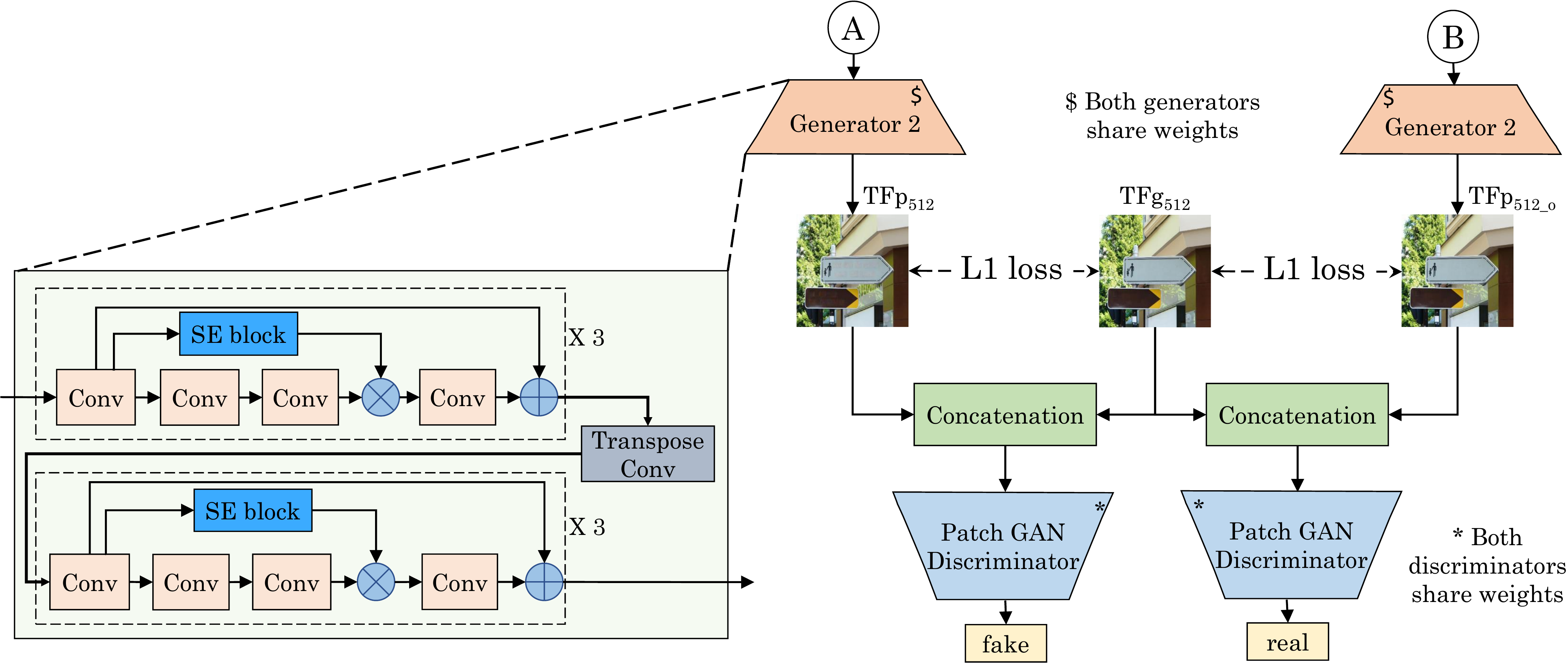}}
 \caption{Image generation (part 2) in TPFNet (\textcircled{\raisebox{-0.9pt}{A}} and \textcircled{\raisebox{-0.9pt}{B}} come from with Part 1 (Figure \ref{fig:stage_1})}
 \label{fig:stage_2}
\end{figure}

\textbf{Q-Block:} The Q-block has two routes (refer to Fig. \ref{fig:ad_block}(a)). The first route incorporates a $1\times1$ convolution, batch normalization, and a ``squeeze and excitation'' block (SE Block). The SE block maps channel dependence and provides access to global information. The second route has two feature-enhancing ConvBlocks. Each block contains a $3\times3$ convolution layer, batch normalization, and GeLU (Gaussian error linear unit) layer.

\textbf{Attention Block:} We employ an attention block, shown in Fig. \ref{fig:ad_block}(b), to achieve high representation ability. The attention block accepts outputs from all three Q-blocks.
The attention block generates attention maps by first concatenating feature maps from branch-1 Q-block and branch-2 Q-block and then applying $1\times1$ convolution and sigmoid activation. Finally, it multiplies this attention map with the result of branch-3 Q-block. The third branch of the subsequent ``M'' module takes its input from the attention block output. Branch-3 benefits from the knowledge acquired by branches 1 and 2. The high-pass filter enhances the feature map's edge-distinguishing capabilities, while the segmentation map pinpoints the exact location of the text.

At last, the fourth module ``M'' produces (1) predicted high-pass filtered image ($Hp_{256}$) (2) predicted segmentation map ($Sp_{256}$) and (3) predicted text-free image ($TFp_{256}$). They are used to train the generator (refer Section \ref{sec:lossFunctionG1}).

\subsection{Discriminator design}\label{sec:discriminator}
In part 1, the discriminator accepts the concatenated vector of text free image and a binary mask showing text position. As a discriminator, the PatchGAN discriminator is used. This discriminator does not attempt to determine whether or not a complete image is real or fraudulent; instead, it analyses and labels small ($N\times N$) patches inside the image.
\subsection{Loss functions used in training}\label{sec:lossFunctionG1}
$G1(.)$ and $D1(.)$ represent the Part-1 generator and discriminator, respectively. $G1(.)$ loss is a combination of four losses: (1) loss for high-pass filtered branch ($H_{loss}$), (2) loss for segmentation branch ($S_{loss}$), (3) loss for text-free image generation branch ($TF_{loss}$) and (4) the conditional adversarial loss between $G1(.)$ and $D1(.)$ ($GAN_{loss}$). Specifically, the overall loss of $G1(.)$ is:
\begin{equation}
  \label{eq5}
  \begin{split}
G1_{loss}& = arg \text{ }\underset{G1}{min}\text{ }\underset{D1}{max}\text{ } GAN_{loss}(G1,D1)+ H_{loss}(Hg_{256},Hp_{256})\\
& + S_{loss}(Sg_{256},Sp_{256}) + TF_{loss}(TFg_{256},TFp_{256})
  \end{split}
\end{equation}

We now describe the individual losses. Note that the ground truth versions of the text-free image, segmentation map and high-pass filtered images are referred to as $TFg_{256}$, $Sg_{256}$, and $Hg_{256}$, respectively.

\textbf{1. Loss for high-pass filtered branch:} The $H_{loss}$ estimated between the ground truth high-pass filtered image and the predicted one is:
\begin{equation}
  \label{eq1}
  H_{loss}(Hg_{256},Hp_{256}) = \sum \left | Hg_{256} - Hp_{256} \right |
\end{equation}

\textbf{2. Loss for segmentation branch:} Here, we use L1 loss and BCE-Dice loss. In the BCE-dice loss, ``binary cross entropy'' (BCE) loss computes the difference in probability distributions between two vectors. The dice loss optimises dice-score results in over-segmented regions. Including the complementary capability of these two losses allows the model to learn more accurate segmentation masks, and the L1 loss regulates outliers. $S_{loss}$ is expressed as:
\begin{equation}
  \label{eq2}
  \begin{split}
  S_{loss}(Sg_{256},Sp_{256}) & = \sum \left | Sg_{256} - Sp_{256} \right | - [\sum Sg_{256} \log (Sp_{256}) \\
  & + (1 - Sg_{256}) \log (1 - Sp_{256})] + (1 - \frac{2 \sum Sg_{256} \times Sp_{256} }{\sum {Sg_{256}}^{2} + \sum {Sp_{256}}^{2}})
  \end{split}
 \end{equation}

\textbf{3. Loss for text-free image generation branch:} $TF_{loss}$ is a combination of L1 loss and SSIM loss. The L1 loss measures the degree to which two images differ in terms of information contained within individual pixels. The SSIM loss accounts for structural details like sharp edges, colour capture, and contrast characteristics. It enhances the similarity index between the ground truth image and the predicted text-free image. $TF_{loss}$ is given as:
\begin{equation}
  \label{eq3}
  \begin{split}
  TF_{loss}(TFg_{256},TFp_{256})& = \sum\left| TFg_{256}- TFp_{256}\right|  +\sum(1- SSIM(TFp_{256}, TFg_{256}))
  \end{split}
\end{equation}

\textbf{4. Loss between $G1(.)$ and $D1(.)$:} Since the segmented mask provides the precise location of the text, we compute $GAN_{loss}$ with the segmented mask as the conditional variable instead of the input image. $GAN_{loss}$ is defined as:
\begin{equation}
  \label{eq4}
  \begin{split}
GAN_{loss}(G1,D1)& = \mathbb{E}_{Sg_{256},TFg_{256}}[\log(D1(Sg_{256},TFg_{256}))]\\
& + \mathbb{E}_{Sp_{256},TFp_{256}} [\log(1 - D1(Sp_{256},TFp_{256}))]
  \end{split}
\end{equation}
\subsection{Part-2: image generation}\label{sec:part2} 

We propose an image generator $G2(.)$, which maps features generated in Part 1 to image space for generating text-free images. We employ a patch GAN image discriminator $D2(.)$ to detect whether an image is real or fake.
In Part 2, the generator concatenates the predicted segmentation map ($Sp_{256}$) and predicted text free image ($TFp_{256}$) vectors to produce the text free image ($TFp_{512}$) with a resolution of $512\times512$. The generator also creates a text-free image ($TFp_{512\_o}$) when provided with a vector consisting of the ground-truth segmentation map ($Sg_{256}$) and the ground-truth text-free image ($TFg_{256}$).

When $G2(.)$ takes input as a concatenated predicted segmentation mask and a predicted text-free image, it generates $TFp_{512}$. When $G2(.)$ takes as input the concatenated ground truth segmentation mask and ground truth text-free image, it generates $TFp_{512\_o}$. We have used the symbol $TFp_{512\_o}$ for the ground-truth concatenated output because it is again used to generate output from generator 2.

The Part-2 generator includes the Conv block. The input is a concatenated vector, which is then fed into a series of three Conv blocks in order to extract features. The output of the first Conv block is then sent through the SE block, and the resulting vector is multiplied by the output of the third Conv block. With the help of the SE block, this draws focus to the crucial aspects. Similarly, the generator employs skip connections to maintain information flow throughout the network. In order to generate a $512\times512$ pixel image without any text, the network utilises transpose convolution to upsample the feature maps, which are run through Conv blocks.

\textbf{Loss function:} The loss is estimated between $TFg_{512}$, $TFp_{512}$ and $TFp_{512\_o}$. The total generator loss of part 2 is:
\begin{equation}
  \label{eq6}
  \begin{split}
G2_{loss} & = arg\text{ }\underset{G2}{min} \text{ }\underset{D2}{max} \text{ } \mathbb{E}_{TFp_{512},TFg_{512}} [\log(D1(TFp_{512},TFg_{512}))]  \\
&+\mathbb{E}_{TFp_{512\_o},TFg_{512}}[\log(1- D1(TFp_{512\_o},TFg_{512}))]\\
& + \sum \left | TFg_{512} - TFp_{512} \right | 
 + \sum \left | TFg_{512} - TFp_{512\_o} \right |
\end{split}
\end{equation}
\section{Experimental Results}
\label{sec:experimental}

\textbf{Implementation settings:} We utilise PyTorch with CUDA 11.2. Experiments are conducted using two A5000 GPUs and a batch size of 32. With an initial learning rate of 1e-4, we train generators using the AdamW optimizer, D1(.) using the RMSprop optimizer, and D2(.) using the Adam optimizer. It is well known that lower-resolution images are simpler to learn from. Since the first part uses a low-resolution image as input, we use RMSprop for the first part discriminator D1(.).
Since part one training can be completed more quickly than part two training, overfitting of the model occurs in part two. However, both parts must be trained concurrently. Thus, using RMSprop for D1(.) causes it to take smaller steps, whereas the use of Adam allows D2(.) in part two to take larger leaps, complementing each other's learning pace. Each part uses the cosine annealing scheduler. We use the same data and hyperparameters for all the baseline training. On 2 RTX A5000 GPUs, training TPFNet took nearly 2.3 days.

\textbf{Metrics:} We use PSNR, SSIM, mean square error (MSE), precision, recall, and F1-Score metrics. PSNR and SSIM help in evaluating the output image quality. For computing the last four metrics, a text detector is used on the output image \cite{nakamura2017scene}. A technique is effective if it achieves a near-zero score on these four metrics.
\subsection{Quantitative Results} 
In this paper, we showcase text removal from the entire image. It is easy to extend our network to remove the text only inside a mask, as shown by Tursun et al. \cite[Figure 1b]{tursun2020mtrnet++}.
The results on Oxford, SCUT-8k, and SCUT-EnsText datasets are shown in Table \ref{tab:oxford_results}, Table \ref{tab:scut_results}, and Table \ref{tab:scut_ensnet_results}, respectively. EnsNet and MTRNet++ were reimplemented with the same settings as TPFNet by replacing our generator with EnsNet's and MTRNet++'s generators, respectively. EnsNet and MTRNet++ were trained on both the SCUT and Oxford datasets in the same manner as TPFNet. Reimplemented results are shown in the table with a (reimplemented) postfix. Other results are taken from previous papers \cite{tursun2020mtrnet++, liu2020erasenet}.

On the Oxford dataset, TPFNet with PVT as backbone gets top PSNR and SSIM values of 44.21 and 0.989, respectively. With EfficientNetB6 as the backbone, we attain the same SSIM value of 0.989 and the second best PSNR value of 37.9. We achieve the least recall value and an F1-Score for text detection, which shows the efficacy of our model. We also achieve the least MSE.

\begin{table}[htbp]
    \centering
    \setlength{\tabcolsep}{2pt}
    \caption{Results on the test set of Oxford Synthetic dataset ($\uparrow$=higher is better, $\downarrow$=lower is better)}
    \label{tab:oxford_results}
    \begin{tabular}{|c|c|c|c|c|c|c|c|}
        \hline
        & Method                             & PSNR $\uparrow$ & SSIM $\uparrow$ & MSE $\downarrow$ & Precision $\downarrow$ & Recall $\downarrow$ & F1-Score $\downarrow$ \\ \hline
        1. & Pix2Pix\cite{isola2017image}       & 24.60 & 0.8970 & 0.54 & 70.03     & 29.34  & 41.35    \\ \hline
        2. & MTRNet\cite{tursun2019mtrnet}     & 29.00 & 0.9300 & 0.20 & \textbf{35.83}     & 0.26   & 0.52     \\ \hline
        3. & EnsNet\cite{zhang2019ensnet}   & 27.40 & 0.9440 & 0.21 & 57.25     & 14.34  & 22.94    \\ \hline
        4. & MTRNet++\cite{tursun2020mtrnet++} & 33.70 & 90.840 & 0.05 & 50.43     & 1.35   & 2.63     \\ \hline
        5. & EnsNet (Reimplemented)         & 28.00 & 0.9790 & 0.24 & 57.21     & 14.32  & 24.11    \\ \hline
        6. & MTRNet++ (Reimplemented)          & 32.99 & 0.9814 & 0.07 & 50.43     & 2.00   & 3.01     \\ \hline
        7. & TPFNet (w/ EfficientNetB6)              & 37.90 & \textbf{0.9890} & 0.01 & 41.00     & 0.12   & 0.21     \\ \hline
        8. & TPFNet (w/ PVT)                 & \textbf{44.21} & \textbf{0.9890} & \textbf{0.01} & 39.00     & \textbf{0.06}   & \textbf{0.17}     \\ \hline
    \end{tabular}    
\end{table}
 
On the SCUT-8K dataset, we obtain a 39.12 PSNR value and a 0.987 SSIM value with PVT as a backbone. With EfficientNetB6 as the backbone, we get the second best result of 36.2 PSNR and 0.973 SSIM. We achieve a 4.52 PSNR percentage point difference between scores of TPFNet and MTRNet++ even though MTRNet++ has used coarse masks. One reason EnsNet findings are comparable to TPFNet on a small-scale dataset is a lack of strong generalization. However, because TPFNet is pre-trained on the Oxford dataset and fine-tuned on the SCUT-8K dataset, it is difficult to overfit even on a small dataset. We achieve an MSE score of 0.01 with both EfficientNetB6 and PVT as the backbone.

\begin{table}[htbp]
    \centering
    \setlength{\tabcolsep}{2pt}
    \caption{Results on the test set of the SCUT-8K dataset}
    \begin{tabular}{|c|c|c|c|c|}
        \hline
           & Method                          & PSNR $\uparrow$ & SSIM $\uparrow$ & MSE $\downarrow$  \\ \hline 
        1. & Pix2Pix\cite{isola2017image}       & 25.60 & 0.4610 & 24.56 \\ \hline 
        2. & STE\cite{nakamura2017scene}    & 14.70 & 0.4610 & 71.48 \\ \hline 
        3. & MTRNet\cite{tursun2019mtrnet}     & 29.70 & 0.9440 & \textbf{0.01}  \\ \hline 
        4. & EnsNet\cite{zhang2019ensnet}   & 37.60 & 0.9640 & 0.20  \\ \hline 
        5. & MTRNet++\cite{tursun2020mtrnet++} & 34.60 & 0.9840 & 0.04  \\ \hline 
        6. & EnsNet (Reimplemented)         & 36.12 & 0.9711 & 0.09  \\ \hline 
        7. & MTRNet++ (Reimplemented)          & 33.67 & 0.9810 & 0.07  \\ \hline 
        8. & TPFNet (w/ EfficientNetB6)              & 36.20 & 0.9730 & \textbf{0.01}  \\ \hline 
        9. & TPFNet (w/ PVT)                 & \textbf{39.12} & \textbf{0.9870} & \textbf{0.01}  \\ \hline 
    \end{tabular}    
    \label{tab:scut_results}
\end{table}

On the SCUT-EnsText dataset, we obtain a PSNR value of 39.0 and an SSIM value of 0.9730 with PVT as the backbone. EraseNet has a close SSIM value of 0.9542, but a higher precision value of 53.20\%, whereas TPFNet has a precision of 21.12\%. We surpass current state-of-the-art approaches in all measures, showing that TPFNet's final output has greater restoration and removal quality.

\begin{table}[htbp]
    \centering
    \setlength{\tabcolsep}{2pt}
    \caption{The results on the test set of the SCUT-ENSTEXT dataset}
    \begin{tabular}{|c|c|c|c|c|c|c|c|}
        \hline
           & Method                          & PSNR $\uparrow$  & SSIM $\uparrow$   & MSE $\downarrow$   & Precision $\downarrow$   & Recall $\downarrow$   & F1score $\downarrow$   \\ \hline
        1. & Pix2Pix\cite{isola2017image}       & 26.69            & 0.8856            & 0.004              & 69.70               & 35.40               & 47.00              \\ \hline
        2. & STE\cite{nakamura2017scene}    & 25.46            & 0.9014            & 0.005              & 40.90               & 5.90                & 10.20              \\ \hline
        3. & EnsNet\cite{zhang2019ensnet}   & 29.53            & 0.9274            & 0.002              & 68.70               & 32.80               & 44.40              \\ \hline
        4. & EraseNet\cite{liu2020erasenet} & 32.29            & 0.9542            & 0.001              & 53.20               & 4.60                & 8.50               \\ \hline
        5 & TPFNet (w/ EfficientNetB6)                 &  37.99   & 0.9700   & 0.00082   & 34.88      & 2.18       & 5.66      \\ \hline
        6 & TPFNet (w/ PVT)                 & \textbf{39.00}   & \textbf{0.9730}   & \textbf{0.00021}   & \textbf{21.12}      & \textbf{1.26}       & \textbf{4.11}      \\ \hline
    \end{tabular}   
    \label{tab:scut_ensnet_results}
\end{table}
   
\subsection{Qualitative Results}
\begin{figure}[htbp]
    \centerline{\includegraphics [width=0.9\textwidth]{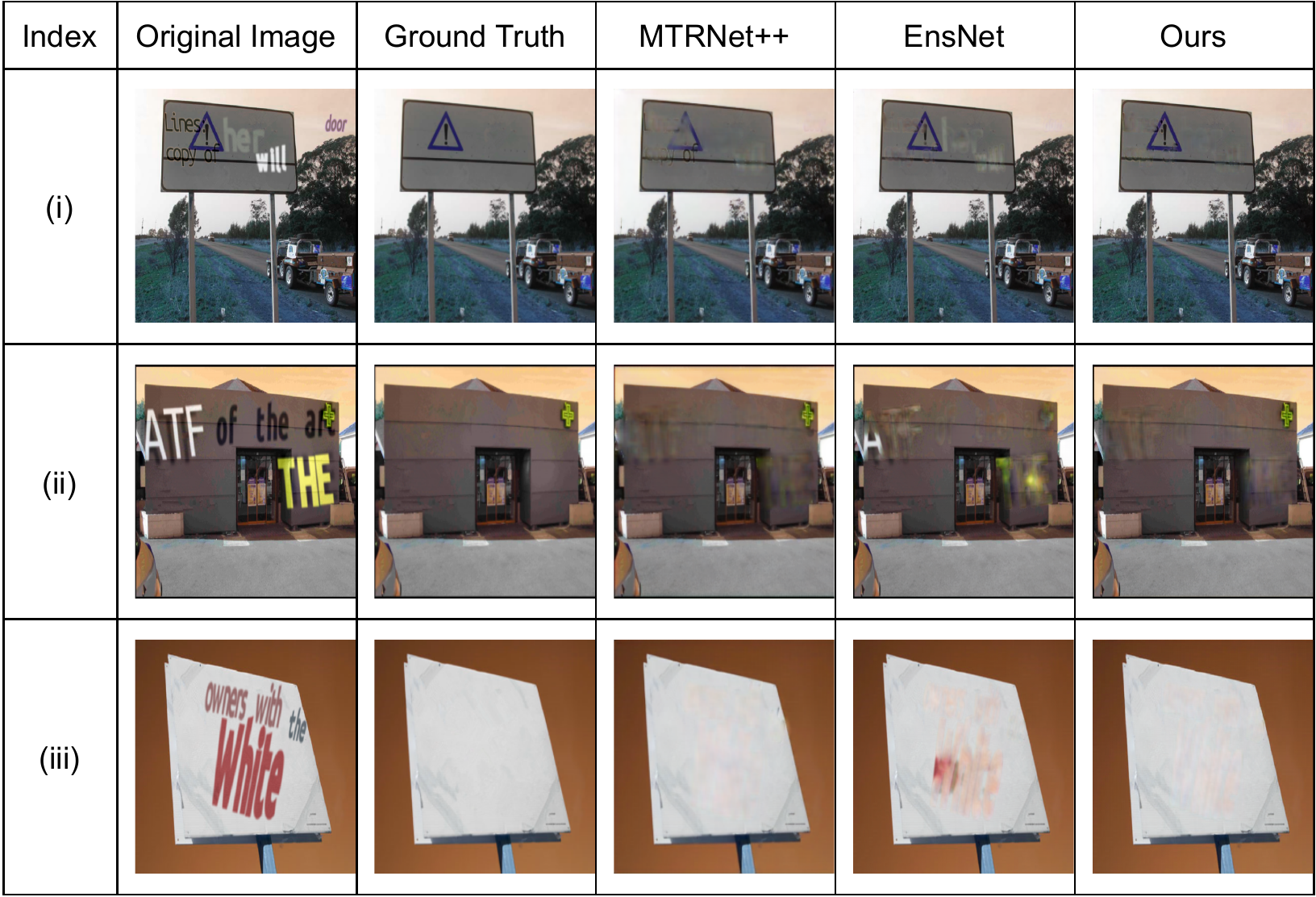}}
    \caption{Qualitative results on SCUT-8K dataset}
    \label{fig:synthetic_results}
\end{figure}

Figure \ref{fig:synthetic_results} shows the qualitative results of the SCUT-8K dataset. Clearly, our results are closer to the ground truth, whereas the outputs of MTRNet++ and EnsNet are blurry. Furthermore, EnsNet's outputs are incomplete and partly corrupted. EnsNet fails to completely remove text from all three images; in fact, it has also removed the plus symbol from row (ii), which is not part of the image text. Similarly, MTRNet++ has removed some portions of the triangle from the row (i), which is not part of the text; it also failed to remove some text. TPFNet is flexible enough to deal with a wide variety of challenging images, including those involving different colours (i), different fonts (ii), and different angles (iii). This shows the robustness of our network in differentiating between text and similar-looking symbols and logos.

\begin{figure}[htbp]
    \centerline{\includegraphics [width=0.9\textwidth]{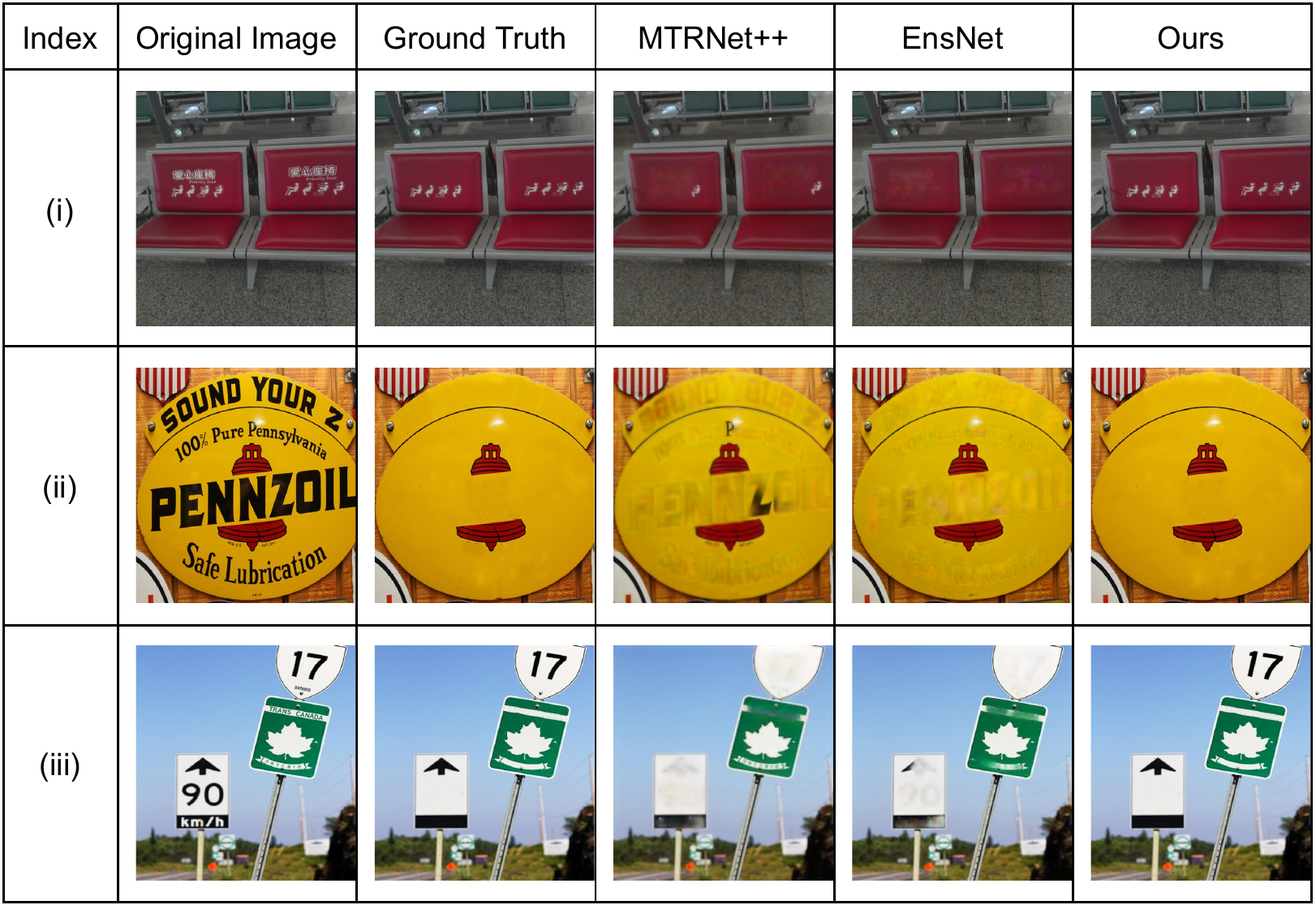}}
    \caption{Qualitative Results on SCUT-EnsText dataset}
    \label{fig:non_synthetic_results}
\end{figure}

Figure \ref{fig:non_synthetic_results} shows the results for the SCUT-EnsText dataset. In row (i), MTRNet++ and EnsNet have obliterated the text; however, they have erroneously also removed similar-looking symbols that are not removed by TPFNet. It can be noticed from row (iii) that MTRNet++ has good performance in text detection over a variety of fonts and text angles; however, it fails to differentiate between text and non-text components properly. TPFNet is effective in detecting the text and differentiating between text and non-text components in all the images.

\section{Ablation Study}
\label{sec:ablation}
Table \ref{tab:ablation} presents the ablation results; we now discuss them.

\textbf{Contribution of branches (S.No. 2 to 4)}: TPFNet has three branches: the high-pass branch, the segmentation branch, and the text-free image generation branch. On removing the high-pass or segmentation branches, PSNR degrades a lot, but SSIM reduces by a smaller amount. Thus, the high-pass branch helps detect edges, and the segmentation branch is helpful for precisely locating the text. The worst results are obtained by removing both high-pass and segmentation branches. Clearly, both branches are essential for effective edge detection with precise text removal.
 
\begin{table}[htbp]
    \centering
    \caption{Ablation results on Scut-EnsText dataset}
    \begin{tabular}{|c|c|c|c|}
    \hline
    & Method                                        & \multicolumn{1}{c|}{PSNR $\uparrow$}  & SSIM $\uparrow$  \\ \hline
    1. & TPFNet                                      & \multicolumn{1}{c|}{39.00} & 0.9730 \\ \hline
    2. & Without high-pass branch                   & \multicolumn{1}{c|}{34.11} & 0.9640 \\ \hline
    3. & Without segmentation branch                & \multicolumn{1}{c|}{32.11} & 0.9631 \\ \hline
    4. & Without high-pass and segmentation branch  & \multicolumn{1}{c|}{30.12} & 0.9412 \\ \hline
    5. & Without Part 2                            & \multicolumn{1}{c|}{37.21} & 0.9701 \\ \hline
    6. & Without pretraning on Oxford dataset       & \multicolumn{1}{c|}{38.23} & 0.9711 \\ \hline    
    \end{tabular}    
    \label{tab:ablation}
\end{table}

\textbf{Contribution of part 2 (S.No. 5):}
Instead of using a $256\times256$ image in the first part, if we directly use a $512\times512$ image in the first part and discard the second part, the quality metrics degrade. This confirms the usefulness of the second part.

\textbf{Benefit from pretraining (S.No. 6):}
We have pre-trained our model on Oxford large-scale dataset and fine-tuned it on the SCUT-8K and SCUT-EnsText datasets. There is a small impact on PSNR and SSIM when we train our network on SCUT-EnsText dataset without pre-training it on Oxford dataset.

\textbf{Impact of backbone:} We changed the backbone from PVT to VGG16 \cite{simonyan2014very}, EfficientNet-B4 \cite{tan2019efficientnet}, ResNet50 \cite{he2016deep}, MobileNetV3-
Large \cite{howard2019searching}, and Swin-Transformer \cite{liu2021swin}. As shown in Table \ref{tab:ablation2}, the best results are achieved by using the Swin-Transformer as the backbone, highlighting the power of the transformer. However, it leads to a very large model size and it does not outperform PVT backbone on other datasets (results omitted); hence, we prefer PVT. EfficientNet-B4 and ResNet50 obtain comparative values, as do VGG16 and MobileNetV3. Evidently, by changing the backbone, a designer can exercise a tradeoff between model size and quality metrics.
 
\begin{table}[htbp]
    \centering
    \caption{Backbone ablation results on Scut-EnsText dataset}
    \begin{tabular}{|c|c|c|c|c|c|}
        \hline
        & Method                                                 & PSNR $\uparrow$ & SSIM $\uparrow$   & \#Parameters (M) $\downarrow$ & FPS $\uparrow$  \\ \hline
        1. & With PVT                                            & 39.00 & 0.9730 & 59.8           & 14    \\ \hline
        2. & With VGG16 \cite{simonyan2014very}                  & 37.23 & 0.9710 & 21.3           & 24    \\ \hline
        3. & With EfficientNet-B4 \cite{tan2019efficientnet}     & 38.41 & 0.9713 & 19.8           & 19    \\ \hline
        4. & With ResNet50 \cite{he2016deep}                     & 38.52 & 0.9723 & 37.9           & 17    \\ \hline
        5. & With MobileNetv3-L \cite{howard2019searching}       & 37.30 & 0.9730 & 12.3           & 26    \\ \hline
        6. & With Swin-Transformer \cite{liu2021swin}            & 39.31 & 0.9763 & 71.8           & 09    \\ \hline
    \end{tabular}
    \label{tab:ablation2}
\end{table}

\vspace{-0.5cm}\section{Conclusion}
\label{sec:conclusion}
In this paper, we proposed an end-to-end deep learning model for text removal from images. Our model outperforms previous work on all the datasets and all the metrics. Our future work will focus on the segmentation of heavily contacted characters in images taken from an oblique angle. Also, we plan to use unsupervised learning approaches to remove text written in languages other than that for which the network was trained.

\bibliographystyle{neurips_2022}  
\bibliography{references}

\end{document}